%% file: main.tex
\documentclass{article}
\usepackage{ijcai22}

\usepackage{times}

\usepackage{soul}
\usepackage{url}
\usepackage[hidelinks]{hyperref}
\usepackage[utf8]{inputenc}
\usepackage[small]{caption}
\usepackage{graphicx}
\usepackage{amsmath}
\usepackage{booktabs} 
\usepackage{multirow}
\usepackage{subfigure}
\usepackage{verbatim}
\usepackage{CJKutf8}
\usepackage{algorithm}
\usepackage{algorithmic}
\usepackage{flushend}

\usepackage{makecell} 
\title{FedCL: Federated Contrastive Learning for Privacy-Preserving Recommendation}

\author{
Chuhan Wu\textsuperscript{\rm $^1$}, 
Fangzhao Wu\textsuperscript{\rm $^2$}, 
Tao Qi\textsuperscript{\rm $^1$}, 
Yongfeng Huang\textsuperscript{\rm $^1$},
Xing Xie\textsuperscript{\rm $^2$}
\affiliations
\textsuperscript{\rm $^1$}Department of Electronic Engineering \& BNRist, Tsinghua University, Beijing 100084, China\\
\textsuperscript{\rm $^2$}Microsoft Research Asia, Beijing 100080, China\\
\emails
  \{wuchuhan15, wufangzhao, taoqi.qt\}@gmail.com, yfhuang@tsinghua.edu.cn, xing.xie@microsoft.com
}

\begin{document}

\maketitle

\begin{abstract}
Contrastive learning is widely used for recommendation model learning, where selecting representative and informative negative samples is critical.
Existing methods usually focus on centralized data, where abundant and high-quality negative samples are easy to obtain.
However, centralized user data storage and exploitation may lead to privacy risks and concerns, while decentralized user data on a single client can be too sparse and biased for accurate contrastive learning.
In this paper, we propose a federated contrastive learning method named \textit{FedCL} for privacy-preserving recommendation, which can exploit high-quality negative samples for effective model training with privacy well protected.
We first infer user embeddings from local user data through the local model on each client, and then perturb them with local differential privacy (LDP) before sending them to a central server for hard negative sampling.
Since individual user embedding contains heavy noise due to LDP, we propose to cluster user embeddings on the server to mitigate the influence of noise, and the cluster centroids are used to retrieve hard negative samples from the item pool.
These hard negative samples are delivered to user clients and mixed with the observed negative samples from local data as well as in-batch negatives constructed from positive samples for federated model training.
Extensive experiments on four benchmark datasets show \textit{FedCL} can empower various recommendation methods in a privacy-preserving way.
\end{abstract}

\input{data/introduction.tex}
\input{data/relatedwork.tex}

\input{data/method.tex}

\input{data/experiment.tex}

\input{data/conclusion.tex}

\bibliographystyle{named}
\bibliography{main}

\appendix
\clearpage

\input{data/supplement}

\end{document}

%% file: data/introduction.tex
\section{Introduction}

Contrastive learning is a widely-adopted technique in personalized recommendation~\cite{zhou2020s3}.
Choosing informative negative samples is usually a necessity in contrastive recommendation model learning~\cite{robinson2020contrastive}, and various strategies have been explored in existing works~\cite{wang2021cross}.
For example, Rendle et al.~\shortcite{rendle2009bpr} proposed to compare each positive sample with a randomly selected negative sample from the entire item set.
Wu et al.~\shortcite{wu2019npa} randomly selected several displayed but not clicked news within a news impression as negative samples to contrast them with the positive sample.
Some recommendation methods further introduce hard negatives to improve the informativeness of training samples.
For instance, Ding et al.~\shortcite{ding2019reinforced} proposed to use reinforcement learning to select hard negative samples from unclicked data.
These methods are usually learned on centralized user data, where abundant and representative negative samples can be drawn from the global dataset.
However, user data is usually highly privacy-sensitive, and centralized storage and mining of it may bring considerable privacy risks and user concerns.
Some strict data protection regulations like GDPR\footnote{https://gdpr-info.eu/} also restrict the collection and centralized storage of user data.
Thus, these methods are not feasible in privacy-preserving recommendation without centralized data to construct training samples.

Recent works study to use federated learning (FL)~\cite{mcmahan2017communication} to learn recommendation models in a decentralized way, where the raw user data is not necessarily uploaded to a server~\cite{chai2020secure}.
For example, Ammad et al.~\shortcite{ammad2019federated} proposed a federated collaborative filtering method that only uploads the item embedding gradients, where negative samples are random items chosen from the item pool.
These methods usually assume that there are sufficient negative items on each local device for negative sampling and model training, which may not be satisfied in real-world scenarios.
In fact, the local data on user devices can be rather sparse and biased, and even there is no negative sample stored on user devices because many online services do not locally record non-interacted items~\cite{ning2021learning}, which poses great challenges to contrastive recommendation model learning under privacy protection.

In this paper, we propose a federated contrastive recommendation model learning method named \textit{FedCL}, which can exploit high-quality negative samples in a privacy-preserving way to empower model training.
In our approach, the local model maintained on each client first infers a user embedding based on local user profiles.
To protect user privacy, we perturb user embeddings via local differential privacy (LDP) before sending them to a central server for hard negative sampling.
Since the perturbed user embeddings are quite noisy, we propose to cluster them on the server via the Ward clustering algorithm to reduce the influence of noise.
The cluster centroids can represent users with different types of interests, and we use the cluster centroids instead of user embeddings to retrieve informative negatives from the item pool via a semi-hard negative sampling method.
These negatives are delivered to user clients for federated model training by mixing them with the real negative samples stored by local clients (if available) and the local in-batch negatives inferred from  positive samples.
Extensive experiments on four benchmark recommendation datasets show that \textit{FedCL} can effectively improve the performance of various recommendation methods and meanwhile protect user privacy.

The main contributions of this paper include: 
\begin{itemize}
    \item We propose a federated contrastive learning method for privacy-preserving recommendation that can exploit informative negatives for model training.
    \item We propose a federated hard negative sampling method that can accurately find hard negative samples for contrastive model training in a privacy-preserving way.
    \item We design a negative mix-up method to incorporate various kinds of negative samples within a unified contrastive training framework.
    \item We conduct extensive experiments to verify the effectiveness of our approach in empowering various recommendation methods in federated learning scenarios.
\end{itemize}

%% file: data/relatedwork.tex
\section{Related Work}\label{sec:RelatedWork}
\subsection{Contrastive Learning in Information Retrieval}

Contrastive learning is a core technique in many information retrieval (IR) tasks such as web search~\cite{xiong2020approximate} and personalized recommendation~\cite{wei2021contrastive}. 
A core problem of contrastive IR model training is selecting proper negative samples to contrast with the positive ones~\cite{robinson2020contrastive}.
One popular way is to randomly sample negatives from the candidate set.
For example, the Bayesian Personalized Ranking (BPR)~\cite{rendle2009bpr} method uses a randomly selected negative sample to compare with a positive sample.
Many other works explore learning IR models by optimizing the gaps between the positive sample and a certain number of random negative samples~\cite{shen2014learning,wu2019npa}.
Another widely used way is in-batch negative sampling.
For example, Yi et al.~\shortcite{yi2019sampling} proposed to use the positive items within the same  batch as shared negatives for all queries in this batch.
However, random or in-batch negative samples can be somewhat too easy for model training~\cite{robinson2020contrastive}.
To exploit more representative and informative negative samples for discriminative model training, some methods explore to incorporate hard negative samples into model learning~\cite{kalantidis2020hard}.
For example, Ding et al.~\shortcite{ding2019reinforced} proposed to use reinforcement learning to train a sampler model to select negative samples for contrastive recommendation model training.
Xiong et al.~\shortcite{xiong2020approximate} proposed to use an asynchronously updated ANN index to select globally hardest negative samples from the candidate document pool for learning the matching model.
These methods usually require centralized data storage for negative sampling.
However, under privacy-preserving settings there is no centralized data and the local data on user devices can be too sparse to obtain effective negative samples, making it difficult to apply  existing contrastive IR model learning methods. 
Thus, in our work, we propose a federated contrastive learning method for recommendation, which can incorporate various kinds of negative samples into contrastive model training in a privacy-preserving way.

\subsection{Federated Recommendation}

In recent years, the application of federated learning in privacy-preserving recommendation has been extensively studied~\cite{chai2020secure,qi2020privacy,liang2021fedrec++}.
For example, Ammad et al.~\shortcite{ammad2019federated} proposed to locally learn user and item embeddings on local user data, and upload item embedding gradients to a server for global model updating and delivery.
Lin et al.~\shortcite{lin2020fedrec} proposed to randomly sample some unrated items and assign them  virtual ratings to avoid communicating the full item embedding table.
Qi et al.~\shortcite{qi2021uni} proposed a unified method for privacy-preserving news recall and ranking by modeling user interests with linear combinations of basis interest vectors.
However, these methods usually assume that there are sufficient negative samples available on local user devices.
Unfortunately, in many scenarios such as product recommendation and web search, non-interacted items are typically not locally recorded due to efficiency concerns.
It may also be suboptimal to simply use random negative items provided by the server without considering clients' characteristics.
One recent work~\cite{ning2021learning} studies learning recommendation models under limited locally available negative samples.
They proposed to use a batch-insensitive loss that enables using local positive items only to construct training samples.
However, useful negative signals cannot be considered by this method.
By contrast, our proposed method can effectively exploit various kinds of negative samples under privacy protection to improve contrastive recommendation model learning.

%% file: data/method.tex
\section{FedCL}

\begin{figure*}[!t]
	\centering 
	 
	\includegraphics[width=0.99\linewidth]{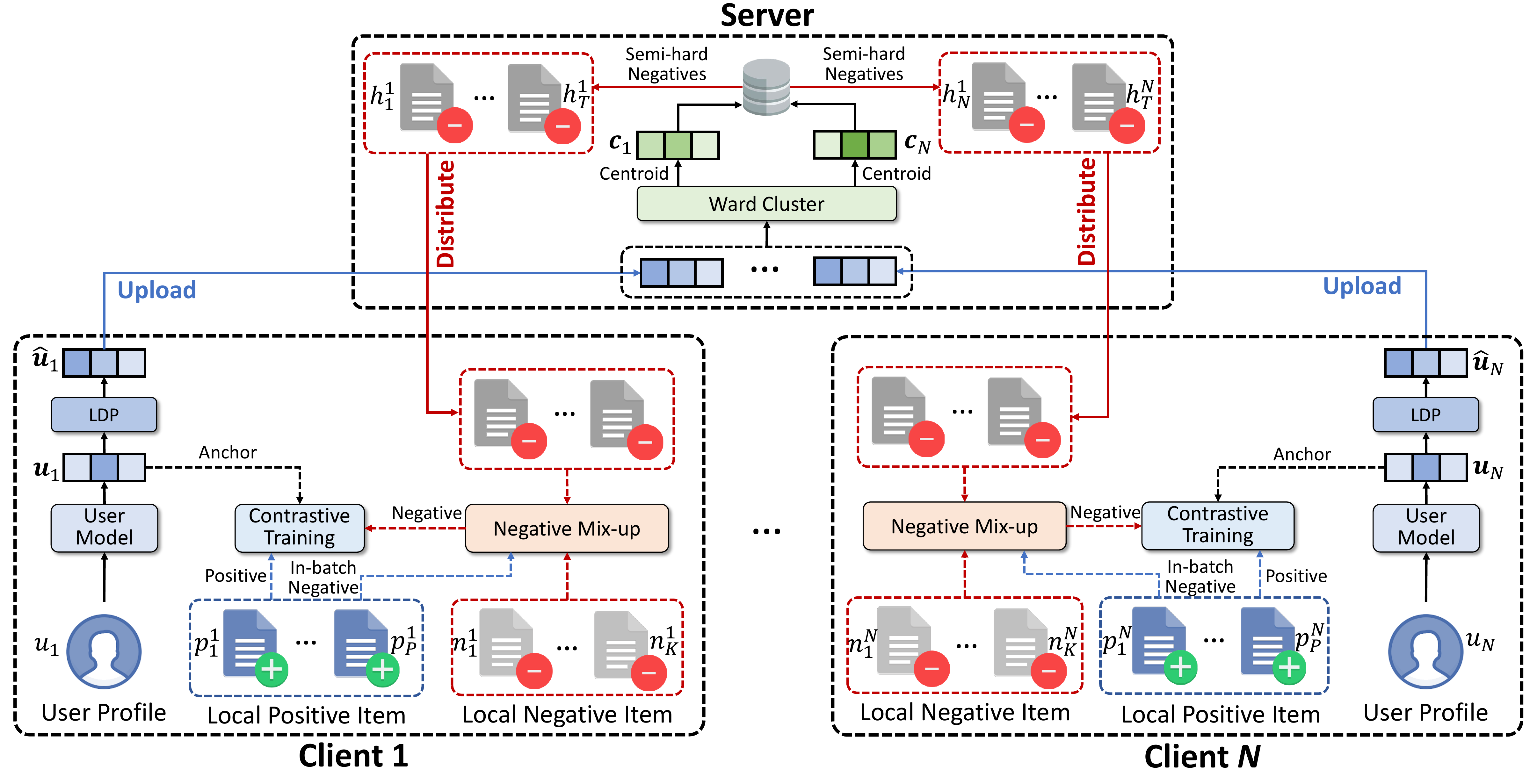} 
\caption{The framework of our \textit{FedCL} approach. The standard local gradients uploading and aggregation in FL are omitted.} \label{fig.model}

\end{figure*}

Next, we introduce our federated contrastive learning  (\textit{FedCL}) approach for recommendation in detail.
Its overall framework is shown in Fig.~\ref{fig.model}.
The server not only coordinates a large number of clients for collaborative model updating, but is also responsible for finding hard negative samples to empower model training.
There are two major steps in \textit{FedCL}, i.e., federated negative sampling and negative mix-up for local contrastive learning, which are described as follows.

\subsection{Federated Negative Sampling}

In federated learning, there is no global dataset for hard negative sampling. 
In addition, since local user data is highly sparse and biased, it is also impractical to conduct hard negative sampling locally.
To solve this problem, we propose a federated negative sampling method to choose hard negatives in a privacy-preserving manner.
Assume that there are $N$ clients in total.
Each client first uses its maintained local user model to infer a user embedding from the local user profile.
The user model can be a simple user ID embedding, or a more complicated one that learns user embeddings from the sequence of historical interacted items.
We denote the embedding of user $u_i$ as $\mathbf{u}_i$.
Since this user embedding still contains much private user information, we use local differential privacy (LDP) to perturb this embedding.
Motivated by~\cite{qi2020privacy}, we first clip each user embedding with a threshold $\delta$ according to its $L_1$ norm, and add Laplace noise $\mathbf{n}\sim Laplace(0, \frac{2\delta}{\epsilon})$ to protect this embedding ($\epsilon$ is the privacy budget, and a lower $\epsilon$ means better privacy protection).
Then the protected user embeddings satisfy $\epsilon$-DP.
We refer to the protected user embedding of $u_i$ as $\mathbf{\hat{u}}_i$.

\begin{figure*}[!t]
	\centering 
	 
	\includegraphics[width=0.75\linewidth]{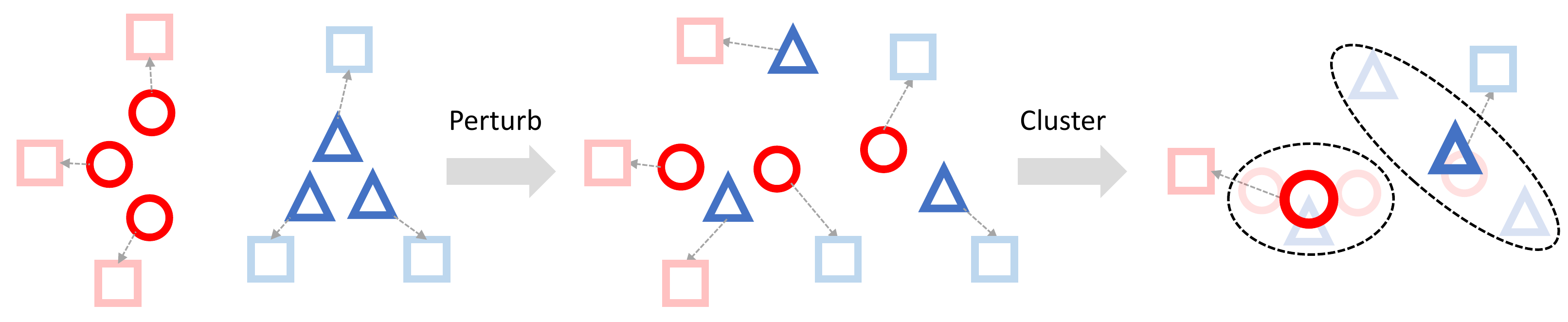} 
\caption{A schematic example of negative sampling based on original user embeddings, perturbed embeddings and cluster centroids. Red circles and blue triangles represent users with two different types of interests. The  squares represent the selected hard negative samples. The perturbed user embeddings cannot select accurate hard negative samples, while using the cluster centroids can reduce the noise impact.} \label{fig.scheme}

\end{figure*}

The protected user embeddings on user clients are uploaded to the server for negative sampling.
A simple way is directly measuring the similarity between candidate items and these embeddings to find negative samples.
However, these embeddings contain heavy noise if the privacy budget is relatively small, and the selected negative samples are usually inaccurate, as shown in the middle plot in Fig.~\ref{fig.scheme}.
To mitigate the influence of noise on negative sampling, we propose to cluster the user embeddings and use their cluster centroids to represent users with different types of interests.
More specifically, we use the Ward clustering algorithm to aggregate the $N$ user embeddings into $C$ clusters.
The centroid of each cluster is the element-wise average of user embeddings in this cluster.
We use the cluster centroid to replace the user embeddings in this cluster to conduct hard negative sampling (right plot in Fig.~\ref{fig.scheme}).
Since the noise added to the user embeddings is zero-mean, the errors brought by the noise can be mitigated to some extent by using cluster centroids.
We denote the cluster centroid corresponding to the user embedding $\mathbf{u}_i$ as $\mathbf{c}_i$.

We evaluate the relevance between $\mathbf{c}_i$ and each item embedding in the candidate pool on the server to choose negative samples.
A simple way is to pick the globally hardest negative samples~\cite{xiong2020approximate}.
However, there may be many false negatives because the candidate items in the entire pool have not  necessarily been displayed to the users.
Thus, in our approach we propose to choose semi-hard negative samples to balance their difficulty and accuracy.
For each centroid vector $\mathbf{c}_i$, we first retrieve $R$\% of candidate samples with the highest difficulties to form a hard candidate subset (a smaller $R$ means harder negatives).
We then randomly select $T$ negative samples from this subset as semi-hard negative samples.
Note that for clients that are associated with the same cluster centroid, the sampling of negative items from the hard candidate subset is independent, which aims to ensure the comprehensiveness of negative samples.
The semi-hard negative samples drawn from $\mathbf{c}_i$ are denoted as $[h_1^i, h_2^i, ..., h_T^i]$, which are further distributed to the  $i$-th client for local model training.
In this way, the user clients can exploit the hard negative samples obtained from the global candidate pool in a privacy-preserving way.

\begin{algorithm}[t]
 \begin{algorithmic}[1]
   \STATE Initialize model parameters on the clients and server
    \REPEAT
    \STATE Select a subset $\mathcal{U}'$ from the user set $\mathcal{U}$ randomly
    \STATE $\mathbf{g}=0$
    \FOR {each user $u_i$ $\in$ $\mathcal{U}'$ in parallel} 
         \STATE Infer user embedding $\mathbf{u}_i$ from local profile
         \STATE $\mathbf{\hat{u}}_i\leftarrow  clip(\mathbf{u}_i,\delta) +Laplace(0,2\delta/\epsilon)$
         \STATE Upload $\mathbf{\hat{u}}_i$ to the server
    \ENDFOR
\\[1.0ex]
    // Federated negative sampling\\[1.0ex]
    
    \STATE Server clusters collected user embeddings
    \FOR {each user $u_i$ $\in$ $\mathcal{U}'$} 
        \STATE Retrieve semi-hard negatives from the item pool using the  cluster centroid $\mathbf{c}_i$ 
        \STATE Send negative items to the client $i$
    \ENDFOR
    \STATE Distribute $\mathbf{g}$ to user clients for local update
\\[1.0ex]
    // Federated Model Learning\\[1.0ex]
    
    \FOR {each user $u_i$ $\in$ $\mathcal{U}'$ in parallel} 
    
        \STATE Mix semi-hard, in-batch and local negatives
    \STATE Compute local loss $\mathcal{L}_i$ on local training samples
    \STATE Compute local model update $\mathbf{g}_i$
    \STATE Securely upload $\mathbf{g}_i$ to the server
    \ENDFOR
    
    \STATE Server aggregates local model updates
    \STATE Update global model and distribute it to clients
       \UNTIL{model convergence} 
     \\[1.0ex]
     
\end{algorithmic}
    \caption{FedCL}
\label{alg}
\end{algorithm}

\subsection{Negative Mix-up for Contrastive Training}

Next, we introduce  the proposed negative mix-up method for local contrastive model learning.
Motivated by the in-batch negative sampling in prior work~\cite{yi2019sampling}, we propose to incorporate in-batch negatives by regarding the local positive items as a mini-batch, where the samples within this batch can be used as negatives for each other.
We denote the local positive samples on the client $i$ as $[p_1^i, p_2^i, ..., p_P^i]$.
For the positive item $p_j^i$, its corresponding in-batch negatives are $[p_1^i, ..., p_{j-1}^i, p_{j+1}..., p_P^i]$.
Note that this kind of in-batch negative samples cannot be used for training ID-based user modeling methods because they cannot learn different user embeddings based on the item interaction histories.
In addition, in real-world scenarios there can be a small number of negative items stored on user devices or randomly distributed by the server.
To help our approach be compatible with this setting, we also incorporate the local negative items into model training, which are denoted as $[n_1^i, n_2^i, ..., n_K^i]$.
Given each positive item $p_j^i$, we mix up the in-batch negatives, local negatives, and semi-hard negatives into a unified list of negative samples, and we denote the list of their embeddings as $\mathcal{N}_j^i$.
The embedding of the positive item $p_j^i$ is denoted as $\mathbf{p}_j^i$
The local contrastive training loss function $\mathcal{L}_i$ of the $i$-th client is formulated as follows:
\begin{equation}
    \mathcal{L}_i=-\sum_{j=1}^P\log(\frac{\exp(\mathbf{u}^i_j\cdot \mathbf{p}_j^i)}{\exp(\mathbf{u}^i_j\cdot \mathbf{p}_j^i)+\sum_{\mathbf{n}\in \mathcal{N}_j^i}\exp(\mathbf{u}^i_j\cdot \mathbf{n})}),
\end{equation}
where $\mathbf{u}^i_j$ represents the user embedding  for matching with the positive item $p_j^i$ (inferred by the user model on client $i$).

\subsection{Federated Model Update}

Finally, we  briefly introduce the federated model update process in \textit{FedCL}.
After computing the loss function $\mathcal{L}_i$, each client locally optimizes this loss to update the maintained local recommendation model.
The local model updates from a certain number of clients are securely uploaded~\cite{chai2020secure} to the server for aggregation, and the aggregated model updates are used to update the global model maintained by the server.
The server further distributes the global model to each client to conduct local update.
The cascaded process of federated negative sampling, local negative mix-up for contrastive training, and federated model update is repeated for multiple rounds until the model converges.

%% file: data/experiment.tex
\section{Experiments}\label{sec:Experiments}

\subsection{Datasets and Experimental Settings}

We conduct experiments on four widely used benchmark datasets for recommendation.
Two of them are taken from the Amazon dataset dump~\cite{mcauley2015image}.
We use the ``Beauty'' and ``Sports and Outdoors'' (shorten as ``Sports'') domains for experiments.
The third dataset is collected on Yelp.
Following the settings in~\cite{zhou2020s3}, we use the transaction records after January 1st, 2019.
The last dataset is MovieLens 1M
(denoted as ML-1M)~\cite{harper2015movielens}, which is also a canonical recommendation dataset.
The statistics of the four datasets are listed in Table~\ref{table.dataset}.
For each user, the last interacted item is used for  test, and the item before the last one is used for validation.

\begin{table}[h]
\centering

\resizebox{0.48\textwidth}{!}{
\begin{tabular}{lrrrrr}
\Xhline{1.0pt}
\multicolumn{1}{c}{Datasets} & \multicolumn{1}{c}{\#Users} & \multicolumn{1}{c}{\#Items} & \multicolumn{1}{c}{\#Actions} & \multicolumn{1}{c}{Avg. length} & \multicolumn{1}{c}{Density} \\ \hline 
Beauty                       & 40,226                      & 54,542                      & 0.35m                         & 8.8                             & 0.02\%                      \\
Sports                        & 25,598	& 18,357      & 0.30m                          & 8.3                            & 0.05\%                      \\
Yelp                         & 30,431                      & 20,033                      & 0.32m                         & 10.4                            & 0.05\%                      \\
ML-1M                        & 6,040                       & 3,416                       & 1.0m                          & 163.5                           & 4.79\%                      \\ 
\Xhline{1.0pt}
\end{tabular}
} 

\caption{Statistics of the four datasets used in experiments.}\label{table.dataset}
\end{table}

\begin{table*}[t]
\centering
\resizebox{0.9\textwidth}{!}{
\begin{tabular}{l|cccc|cccc}
\Xhline{1.0pt}
\multicolumn{1}{c|}{\multirow{2}{*}{\textbf{Methods}}} & \multicolumn{4}{c|}{\textbf{Beauty}} & \multicolumn{4}{c}{\textbf{Sports}} \\ \cline{2-9} 
\multicolumn{1}{c|}{}                                  & HR@5    & HR@10  & nDCG@5 & nDCG@10 & HR@5   & HR@10  & nDCG@5 & nDCG@10 \\ \hline
NCF                                                   & 0.0144          & 0.0296          & 0.0082          & 0.0131          & 0.0124          & 0.0220          & 0.0068          & 0.0101          \\
NCF+Fed                                               & 0.0122          & 0.0267          & 0.0070          & 0.0123          & 0.0106          & 0.0199          & 0.0062          & 0.0094          \\
NCF+BIL                                               & 0.0140          & 0.0288          & 0.0079          & 0.0129          & 0.0125          & 0.0223          & 0.0070          & 0.0102          \\
NCF+FedCL                                             & 0.0245          & 0.0413          & 0.0155          & 0.0209          & 0.0178          & 0.0304          & 0.0109          & 0.0148          \\ \hline
GRU4Rec                                               & 0.0166          & 0.0358          & 0.0088          & 0.0143          & 0.0135          & 0.0277          & 0.0094          & 0.0135          \\
GRU4Rec+Fed                                           & 0.0134          & 0.0316          & 0.0074          & 0.0130          & 0.0115          & 0.0227          & 0.0079          & 0.0104          \\
GRU4Rec+BIL                                           & 0.0160          & 0.0343          & 0.0083          & 0.0136          & 0.0130          & 0.0270          & 0.0091          & 0.0129          \\
GRU4Rec+FedCL                                         & 0.0312          & 0.0539          & 0.0189          & 0.0265          & 0.0184          & 0.0311          & 0.0115          & 0.0156          \\ \hline
BERT4Rec                                              & 0.0199          & 0.0410          & 0.0188          & 0.0259          & 0.0169          & 0.0332          & 0.0104          & 0.0156          \\
BERT4Rec+Fed                                          & 0.0158          & 0.0352          & 0.0160          & 0.0222          & 0.0134          & 0.0292          & 0.0098          & 0.0135          \\
BERT4Rec+BIL                                          & 0.0191          & 0.0397          & 0.0182          & 0.0250          & 0.0164          & 0.0325          & 0.0101          & 0.0150          \\
BERT4Rec+FedCL                                        & \textbf{0.0368} & \textbf{0.0654} & \textbf{0.0209} & \textbf{0.0304} & \textbf{0.0215} & \textbf{0.0348} & \textbf{0.0124} & \textbf{0.0179} \\ \Xhline{1.0pt}
\end{tabular}
}
\caption{Results of different methods on the Beauty and Sports datasets.}\label{table2}
\end{table*}

\begin{table*}[t]
\centering
\resizebox{0.9\textwidth}{!}{
\begin{tabular}{l|cccc|cccc}
\Xhline{1.0pt}
\multicolumn{1}{c|}{\multirow{2}{*}{\textbf{Methods}}} & \multicolumn{4}{c}{\textbf{Yelp}} & \multicolumn{4}{c}{\textbf{ML-1M}} \\ \cline{2-9} 
\multicolumn{1}{c|}{}                                  & HR@5    & HR@10  & nDCG@5 & nDCG@10 & HR@5   & HR@10  & nDCG@5 & nDCG@10 \\ \hline
NCF                                                   & 0.0134 & 0.0275 & 0.0080 & 0.0118  & 0.0168 & 0.0322 & 0.0098 & 0.0145  \\
NCF+Fed                                               & 0.0104 & 0.0212 & 0.0072 & 0.0109  & 0.0133 & 0.0275 & 0.0090 & 0.0137  \\
NCF+BIL                                               & 0.0128 & 0.0266 & 0.0078 & 0.0116  & 0.0157 & 0.0310 & 0.0095 & 0.0142  \\
NCF+FedCL                                             & 0.0213 & 0.0359 & 0.0135 & 0.0186  & 0.0414 & 0.0675 & 0.0256 & 0.0342  \\ \hline
GRU4Rec                                               & 0.0153 & 0.0262 & 0.0101 & 0.0135  & 0.0769 & 0.1662 & 0.0388 & 0.0674  \\
GRU4Rec+Fed                                           & 0.0120 & 0.0237 & 0.0084 & 0.0112  & 0.0588 & 0.1025 & 0.0309 & 0.0531  \\
GRU4Rec+BIL                                           & 0.0146 & 0.0255 & 0.0098 & 0.0129  & 0.0754 & 0.1633 & 0.0372 & 0.0660  \\
GRU4Rec+FedCL                                         & 0.0264 & 0.0413 & 0.0176 & 0.0225  & \textbf{0.1155} & \textbf{0.2013} & \textbf{0.0671} & \textbf{0.0940}  \\ \hline
BERT4Rec                                              & 0.0184 & 0.0292 & 0.0117 & 0.0173  & 0.0736 & 0.1344 & 0.0427 & 0.0622  \\
BERT4Rec+Fed                                          & 0.0152 & 0.0256 & 0.0100 & 0.0114  & 0.0544 & 0.0939 & 0.0288 & 0.0487  \\
BERT4Rec+BIL                                          & 0.0178 & 0.0299 & 0.0119 & 0.0180  & 0.0727 & 0.1316 & 0.0420 & 0.0616  \\
BERT4Rec+FedCL                                        & \textbf{0.0281} & \textbf{0.0440} & \textbf{0.0189} & \textbf{0.0236}  & 0.1115 & 0.1946 & 0.0642 & 0.0909  \\ \Xhline{1.0pt}
\end{tabular}
}
\caption{Results of different methods on the Yelp and ML-1M datasets.}\label{table3}
\end{table*}

We use the Adam~\cite{kingma2014adam} optimizer for model training.
The learning rate is 1e-3, and the number of selected users per update is 16.
The hidden dimension is 64.
The user embedding norm clip threshold is 1.
The privacy budget $\epsilon$ is 4.
The number of clusters is 25.
To simulate local negative items, we randomly sample 100 items on each client, and for each positive we randomly choose 10 local negatives to participate in contrastive learning.
The difficulty $R$\% of negative samples is set to 25\%.
The number of semi-hard negatives is 20.
These hyperparameters are tuned on the validation sets.
Motivated by~\cite{krichene2020sampled}, we evaluate  the recommendation performance on the whole item set rather than the sampled subsets used in~\cite{sun2019bert4rec}.
We use the Hit Ratio (HR) and normalized Discounted Cumulative Gain (nDCG) metrics over the top 5 or 10 ranked items.
We repeat each experiment 5 times and report the average scores.
We will soon release our code for reproducibility.

\subsection{Performance Evaluation}

To verify the effectiveness and generality of \textit{FedCL}, we compare the results of several widely used methods, including NCF~\cite{he2017neural}, GRU4Rec~\cite{hidasi2015session} and BERT4Rec~\cite{sun2019bert4rec}, under four different settings as follows:
(a) centralized learning, which uses the original data sampling and model learning methods based on centralized data;
(b) federated learning (Fed), which directly applies the federated learning framework to the basic model (local negatives are used);
(c) federated learning with batch-insensitive loss (BIL)~\cite{ning2021learning}, which uses batch-insensitive loss to learn models on local positive items only;
(d) federated contrastive learning (FedCL), our proposed method.
The results on the four datasets are shown in Tables~\ref{table2} and~\ref{table3}.
We find that compared with centralized learning, the performance of models learned in the standard federated framework usually decreases.
This is mainly due to the non-IID property of user data decentralized on different clients and the scarcity of informative negative samples.
In addition, the \textit{BIL} method outperforms the federated baseline.
This is because BIL learns uniform item representations via a spreadout regularization instead of learning from the limited local negative samples.
Moreover, our proposed \textit{FedCL} method outperforms other methods (t-test results show the improvement is significant, $p<0.001$), and it can consistently boost the performance of different base models.
This is because our approach can incorporate various kinds of negative samples into  contrastive model training to exploit richer negative signals.

\subsection{Influence of Negative Samples}

\begin{figure}[!t]
	\centering 
	 
	\includegraphics[width=0.99\linewidth]{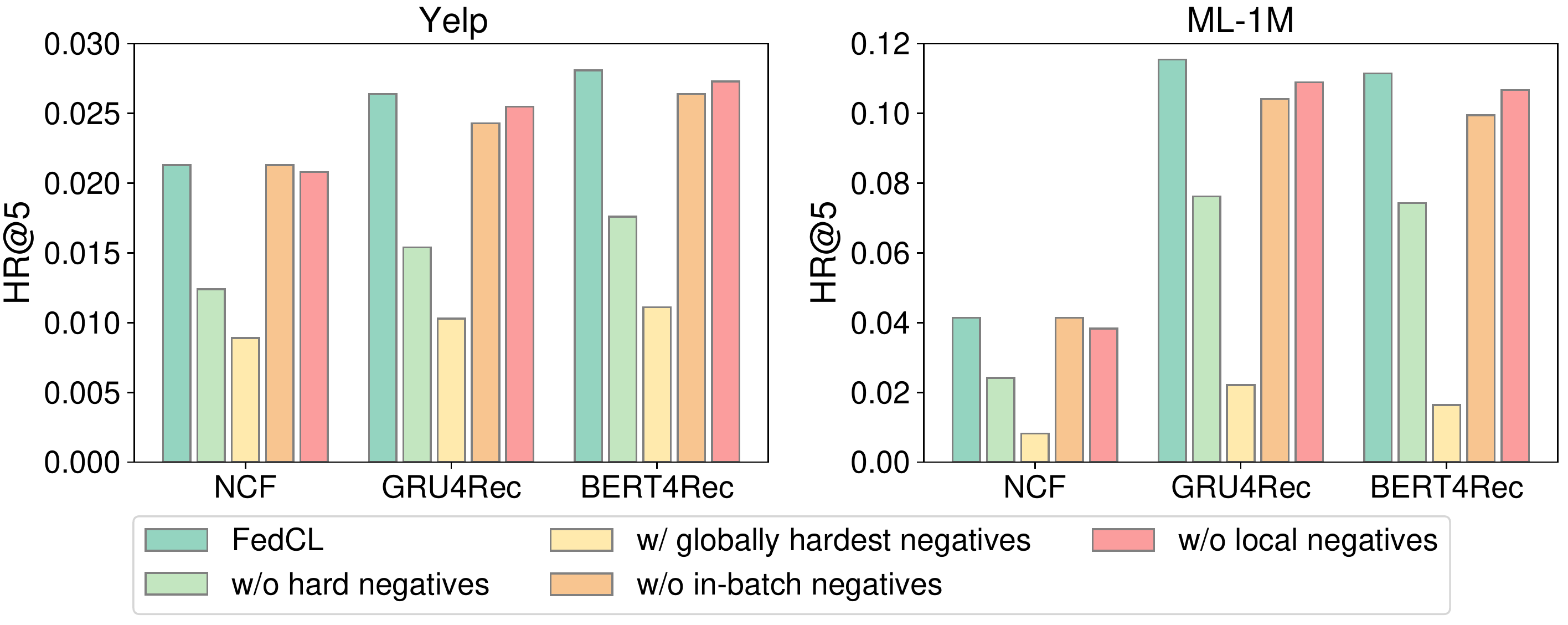} 
\caption{Influence of different types of negative samples.} \label{fig.exp1}

\end{figure}

Next, we study the impacts of different types of negative samples in our approach.
We compare the model performance with one kind of negative samples removed, and the performance of replacing semi-hard negatives with globally hardest negatives~\cite{xiong2020approximate}.
The results on the Yelp and ML-1M datasets are shown in Fig.~\ref{fig.exp1}.
We find that the hard negatives obtained by the federated negative sampling method have a major contribution to model performance.
However, the performance has a huge decline if they are replaced by globally hardest samples.
This is mainly because globally hardest samples may contain many false negatives, which are harmful to model training.
In addition, both in-batch and local negatives have  contributions, but the performance gain brought by the local negatives is not so large.
Thus, the performance of \textit{FedCL} can still be promising if there is no local negative sample available on user devices.

\subsection{Influence of Clustering}

\begin{figure}[!t]
	\centering 
	 
	\includegraphics[width=0.99\linewidth]{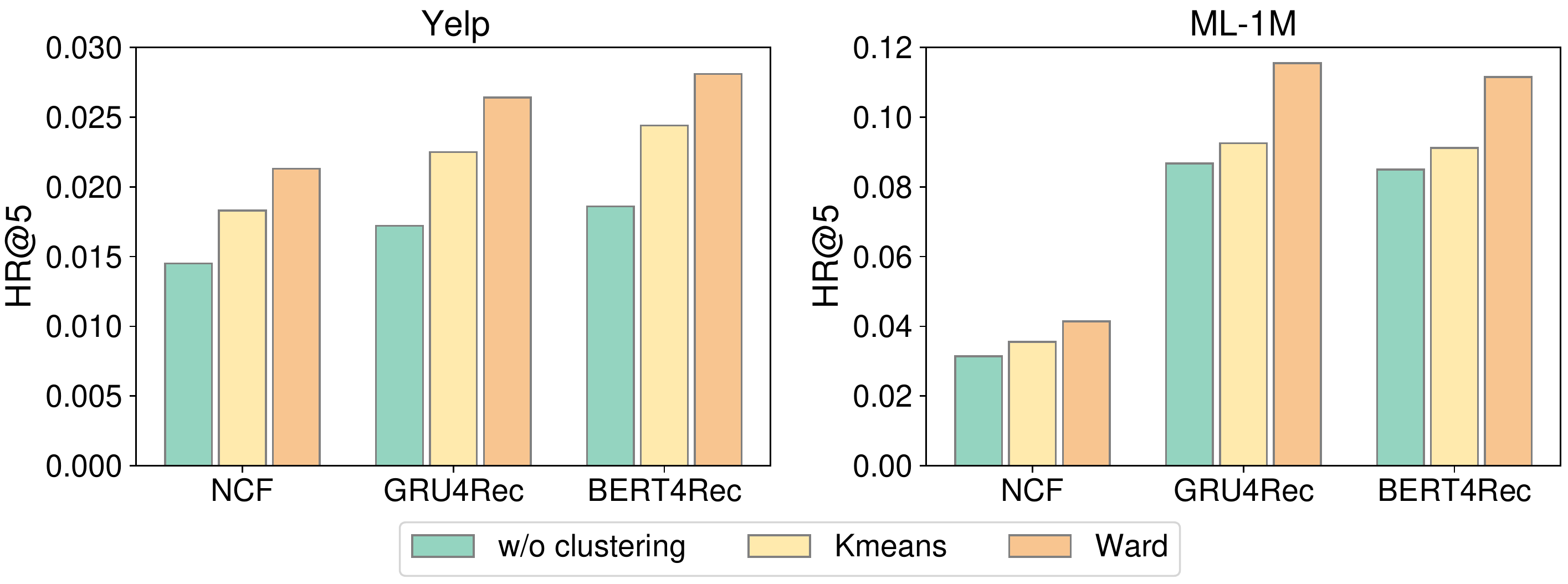} 
\caption{Influence of user embedding clustering.} \label{fig.exp2}

\end{figure}

We further study the influence of clustering on the model performance.
The results without clustering, with Kmeans clustering, or with Ward clustering are compared in Fig.~\ref{fig.exp2}.
We observe that the performance is suboptimal when the user embeddings are directly used for hard negative sampling.
This is intuitive because the perturbed user embeddings are quite noisy.
In addition, both Kmeans and Ward clustering algorithms can improve the model performance.
This may be because using the centroids of clusters to replace original user embeddings can mitigate the influence of noise.
In addition, the Ward clustering method is better than Kmeans.
This may be because the hierarchical clustering method is stronger in modeling the relations between similar user interests.

\subsection{Influence of Cluster Number}
\begin{figure}[!t]
	\centering 
	 
	\includegraphics[width=0.99\linewidth]{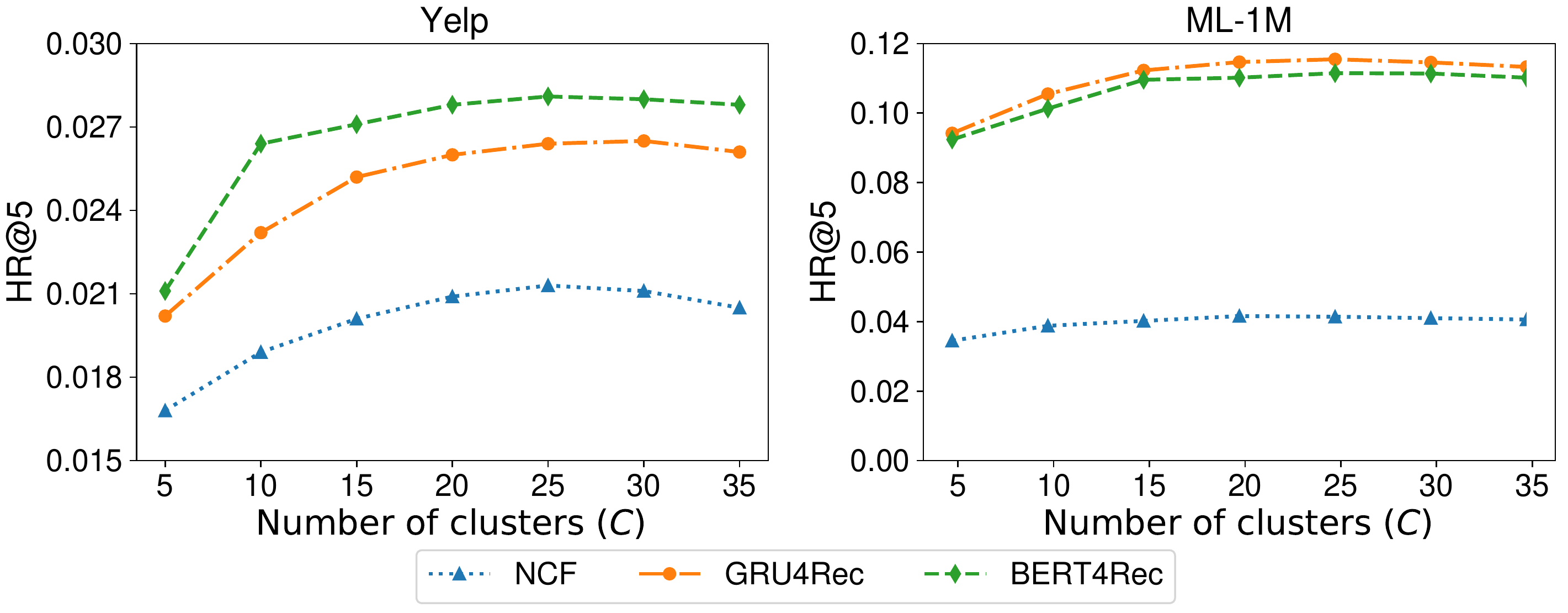} 
\caption{Influence of the number of clusters.} \label{fig.exp3}

\end{figure}

We then analyze the impacts of the number of clusters on model performance, as shown in Fig.~\ref{fig.exp3}.
The results reveal that when the number of clusters is too small, the performance is suboptimal.
This may be because user interests cannot be effectively distinguished when there are too few clusters.
However, the performance starts to decrease when the number of clusters is larger than 25.
This may be because the influence of noise encoded in user embeddings becomes large.
Thus, we set the number of clusters $C$ to 25.

\subsection{Influence of Privacy Budget}

\begin{figure}[!t]
	\centering 
	 
	\includegraphics[width=0.99\linewidth]{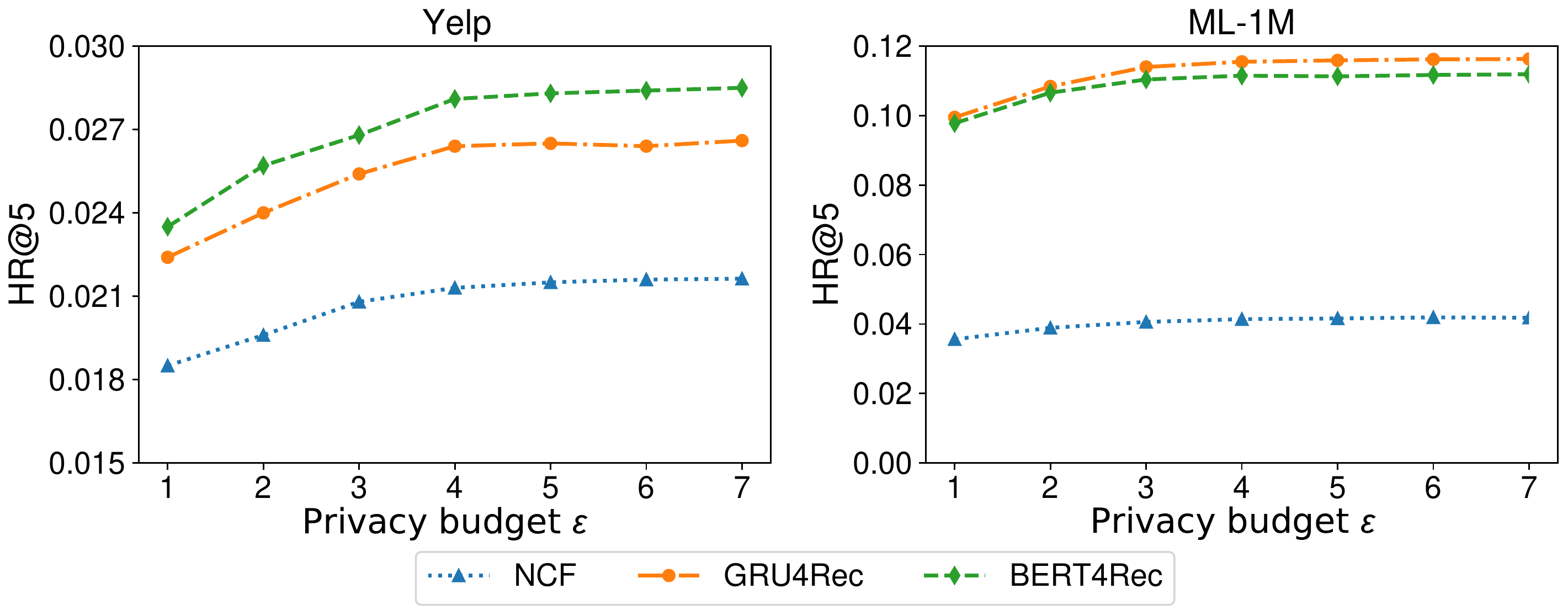} 
\caption{Influence of the privacy budget of user embeddings for hard negative sampling.} \label{fig.exp4}

\end{figure}

Finally, we analyze the influence of the privacy budget of user embeddings for hard negative sampling on the model performance.
We change the intensity of Laplace noise according to different privacy budgets, and the corresponding results are shown in Fig.~\ref{fig.exp4}.
We find that the performance usually decreases when a smaller privacy budget $\epsilon$ is used.
This is intuitive because the noise is stronger when $\epsilon$ is smaller, and the obtained negative samples are more inaccurate.
In our approach, we choose $\epsilon=4$ that can yield a good performance and satisfactory user privacy protection ability.

\subsection{Experiments in Appendix}

In the appendix, we analyze the impact of the negative sample difficulty ($R$\%) as well as the number of local and semi-hard negative samples.
The results show that selecting negatives from the top 25\% relevant candidates and using a medium number of negative samples is suitable.

%% file: data/conclusion.tex
\section{Conclusion}\label{sec:Conclusion}

In this paper, we propose a federated contrastive learning framework for privacy-preserving recommendation, which can effectively exploit various kinds of informative negative samples to empower federated recommendation model learning.
In our approach, we propose a federated negative sampling method to choose semi-hard negatives from the candidate item pool in a privacy-preserving way based on the locally inferred and perturbed user embeddings.
In addition, we explore using a negative mix-up method to jointly incorporate local, in-batch and semi-hard negative samples into a unified contrastive learning framework.
Extensive experiments on four widely used benchmark recommendation datasets validate that our approach can effectively improve different recommendation methods in a privacy-preserving manner.

%% file: data/supplement.tex
\section*{Appendix}

\subsection*{Implementation Details}
Several implementation details are introduced as follows.
We conducted experiments using a machine with Ubuntu 16.04 operating system and Python 3.6.
The machine has a memory of 256GB and a Tesla V100 GPU with 32GB memory.
We used Keras 2.2.4 and tensorflow 1.15 to implement deep learning models.
The Faiss library is used to retrieve negative samples.

\subsection*{Influence of Negative Sample Difficulty}

\begin{figure}[h]
	\centering  
	\includegraphics[width=0.48\textwidth]{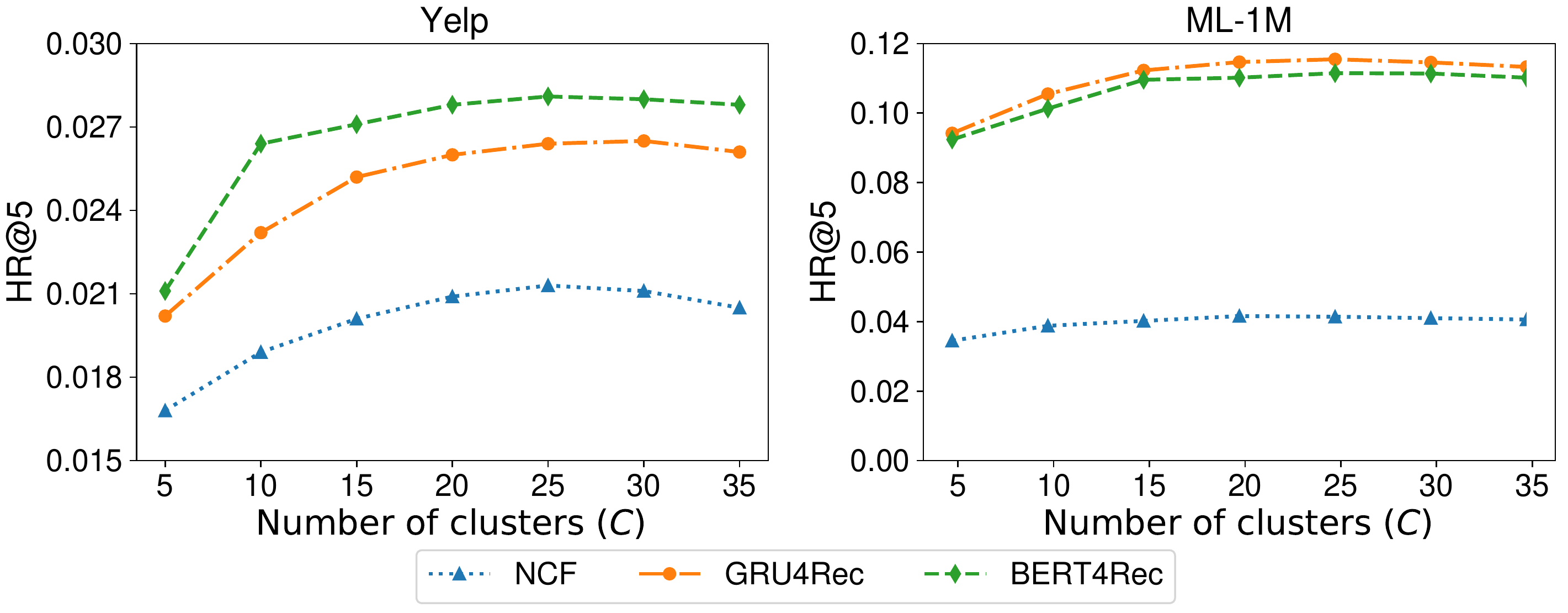}  
\caption{Influence of the negative sample difficulty. } \label{fig.exps1}
\end{figure}

Fig.~\ref{fig.exps1} shows the influence of using  top $R$\% relevant candidate items for semi-hard negative sampling.
The results show that the performance is suboptimal when the ratio $R$ is either too small or too large.
This is because when $R$ is too large the sampled items are not so informative,  while a small $R$ may lead to inaccurate negatives because they are too hard.
Thus, we choose $R=25$ in our experiments.

\subsection*{Influence of Sample Number}
 
\begin{figure}[h]
	\centering  
	\includegraphics[width=0.48\textwidth]{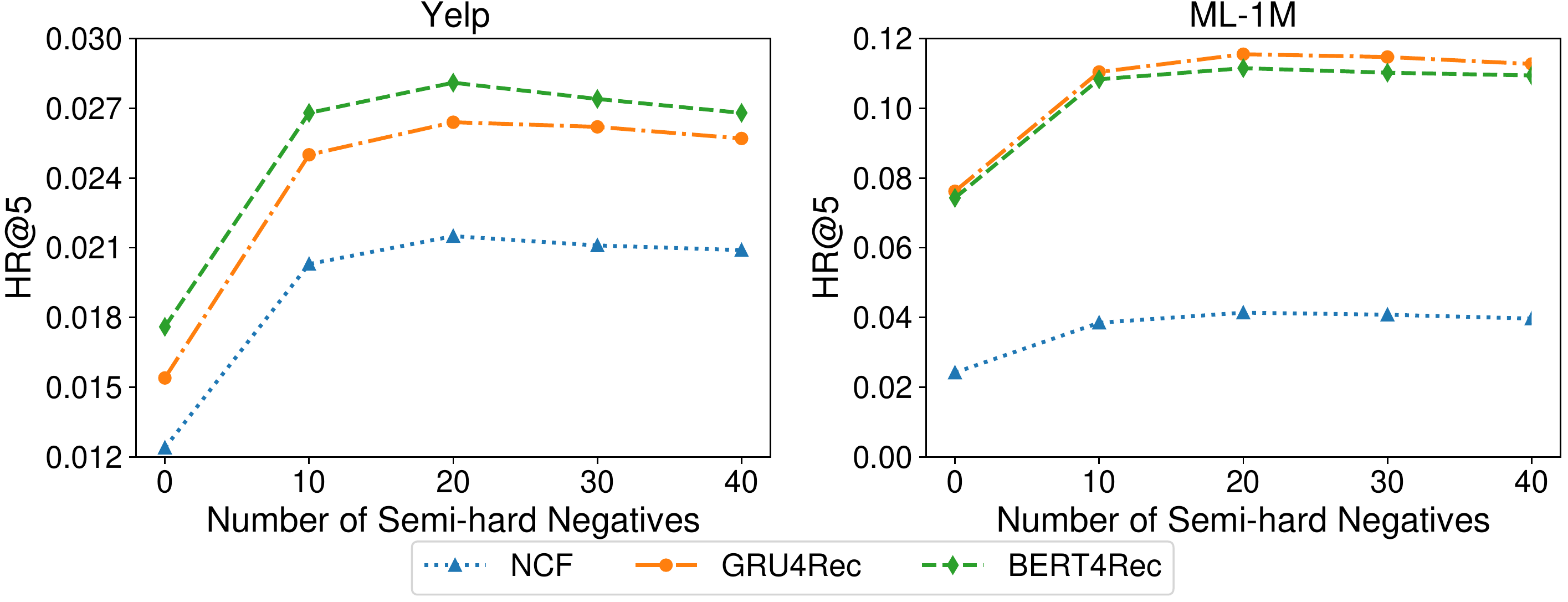}  
\caption{Influence of the number of semi-hard negative samples. } \label{fig.exps2}
\end{figure}

We also study the influence of semi-hard and local negative samples on the model performance.
We first set the number of local negative samples to 10 and change the number of semi-hard negative samples, as shown in Fig.~\ref{fig.exps2}.
We find the performance is optimal when about 20 semi-hard negative samples are used.
This is because the negative signals cannot be effectively exploited when semi-hard negatives are insufficient, while the negative sets are not uniform enough when there are too many hard negatives.

\begin{figure}[h]
	\centering  
	\includegraphics[width=0.48\textwidth]{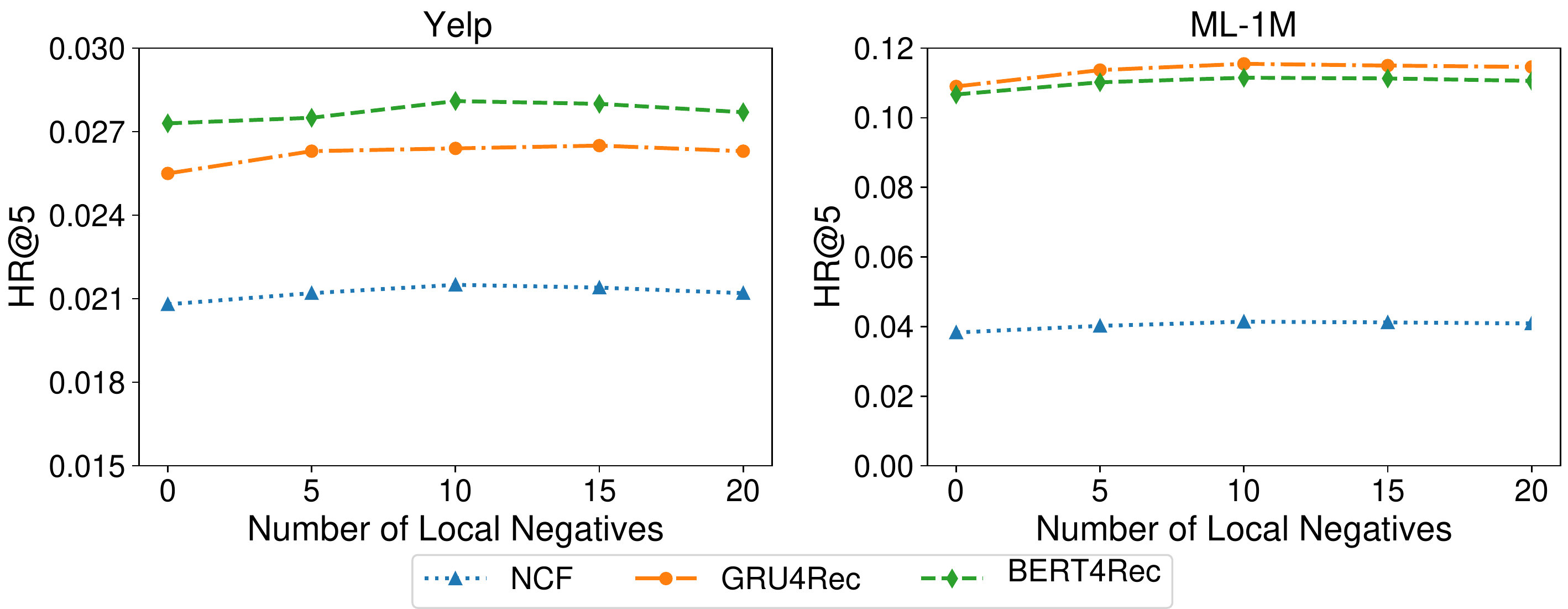}  
\caption{Influence of the number of local negative samples. } \label{fig.exps3}
\end{figure}

We further change the number of local negative samples and compare the results (Fig.~\ref{fig.exps3}).
We find that the number of local negative samples used for model training also needs to be moderate.
Thus, we choose to use 10 negative samples in model learning.


%% file: main.bbl
\begin{thebibliography}{}

\bibitem[\protect\citeauthoryear{Ammad-Ud-Din \bgroup \em et al.\egroup
  }{2019}]{ammad2019federated}
Muhammad Ammad-Ud-Din, Elena Ivannikova, Suleiman~A Khan, Were Oyomno, Qiang
  Fu, Kuan~Eeik Tan, and Adrian Flanagan.
\newblock Federated collaborative filtering for privacy-preserving personalized
  recommendation system.
\newblock {\em arXiv preprint arXiv:1901.09888}, 2019.

\bibitem[\protect\citeauthoryear{Chai \bgroup \em et al.\egroup
  }{2020}]{chai2020secure}
Di~Chai, Leye Wang, Kai Chen, and Qiang Yang.
\newblock Secure federated matrix factorization.
\newblock {\em IEEE Intelligent Systems}, 2020.

\bibitem[\protect\citeauthoryear{Ding \bgroup \em et al.\egroup
  }{2019}]{ding2019reinforced}
Jingtao Ding, Yuhan Quan, Xiangnan He, Yong Li, and Depeng Jin.
\newblock Reinforced negative sampling for recommendation with exposure data.
\newblock In {\em IJCAI}, pages 2230--2236, 2019.

\bibitem[\protect\citeauthoryear{Harper and
  Konstan}{2015}]{harper2015movielens}
F~Maxwell Harper and Joseph~A Konstan.
\newblock The movielens datasets: History and context.
\newblock {\em TIIS}, 5(4):1--19, 2015.

\bibitem[\protect\citeauthoryear{He \bgroup \em et al.\egroup
  }{2017}]{he2017neural}
Xiangnan He, Lizi Liao, Hanwang Zhang, Liqiang Nie, Xia Hu, and Tat-Seng Chua.
\newblock Neural collaborative filtering.
\newblock In {\em WWW}, pages 173--182, 2017.

\bibitem[\protect\citeauthoryear{Hidasi \bgroup \em et al.\egroup
  }{2016}]{hidasi2015session}
Bal{\'a}zs Hidasi, Alexandros Karatzoglou, Linas Baltrunas, and Domonkos Tikk.
\newblock Session-based recommendations with recurrent neural networks.
\newblock In {\em ICLR}, 2016.

\bibitem[\protect\citeauthoryear{Kalantidis \bgroup \em et al.\egroup
  }{2020}]{kalantidis2020hard}
Yannis Kalantidis, Mert~Bulent Sariyildiz, Noe Pion, Philippe Weinzaepfel, and
  Diane Larlus.
\newblock Hard negative mixing for contrastive learning.
\newblock {\em arXiv preprint arXiv:2010.01028}, 2020.

\bibitem[\protect\citeauthoryear{Kingma and Ba}{2015}]{kingma2014adam}
Diederik~P. Kingma and Jimmy Ba.
\newblock Adam: A method for stochastic optimization.
\newblock In {\em ICLR}, 2015.

\bibitem[\protect\citeauthoryear{Krichene and
  Rendle}{2020}]{krichene2020sampled}
Walid Krichene and Steffen Rendle.
\newblock On sampled metrics for item recommendation.
\newblock In {\em KDD}, pages 1748--1757, 2020.

\bibitem[\protect\citeauthoryear{Liang \bgroup \em et al.\egroup
  }{2021}]{liang2021fedrec++}
Feng Liang, Weike Pan, and Zhong Ming.
\newblock Fedrec++: Lossless federated recommendation with explicit feedback.
\newblock In {\em AAAI}, pages 4224--4231, 2021.

\bibitem[\protect\citeauthoryear{Lin \bgroup \em et al.\egroup
  }{2020}]{lin2020fedrec}
Guanyu Lin, Feng Liang, Weike Pan, and Zhong Ming.
\newblock Fedrec: Federated recommendation with explicit feedback.
\newblock {\em IEEE Intelligent Systems}, 2020.

\bibitem[\protect\citeauthoryear{McAuley \bgroup \em et al.\egroup
  }{2015}]{mcauley2015image}
Julian McAuley, Christopher Targett, Qinfeng Shi, and Anton Van Den~Hengel.
\newblock Image-based recommendations on styles and substitutes.
\newblock In {\em SIGIR}, pages 43--52, 2015.

\bibitem[\protect\citeauthoryear{McMahan \bgroup \em et al.\egroup
  }{2017}]{mcmahan2017communication}
Brendan McMahan, Eider Moore, Daniel Ramage, Seth Hampson, and Blaise~Aguera
  y~Arcas.
\newblock Communication-efficient learning of deep networks from decentralized
  data.
\newblock In {\em AISTATS}, pages 1273--1282. PMLR, 2017.

\bibitem[\protect\citeauthoryear{Ning \bgroup \em et al.\egroup
  }{2021}]{ning2021learning}
Lin Ning, Karan Singhal, Ellie~X Zhou, and Sushant Prakash.
\newblock Learning federated representations and recommendations with limited
  negatives.
\newblock {\em arXiv preprint arXiv:2108.07931}, 2021.

\bibitem[\protect\citeauthoryear{Qi \bgroup \em et al.\egroup
  }{2020}]{qi2020privacy}
Tao Qi, Fangzhao Wu, Chuhan Wu, Yongfeng Huang, and Xing Xie.
\newblock Privacy-preserving news recommendation model learning.
\newblock In {\em EMNLP: Findings}, pages 1423--1432, 2020.

\bibitem[\protect\citeauthoryear{Qi \bgroup \em et al.\egroup
  }{2021}]{qi2021uni}
Tao Qi, Fangzhao Wu, Chuhan Wu, Yongfeng Huang, and Xing Xie.
\newblock Uni-fedrec: A unified privacy-preserving news recommendation
  framework for model training and online serving.
\newblock In {\em EMNLP: Findings}, pages 1438--1448, 2021.

\bibitem[\protect\citeauthoryear{Rendle \bgroup \em et al.\egroup
  }{2009}]{rendle2009bpr}
Steffen Rendle, Christoph Freudenthaler, Zeno Gantner, and Lars Schmidt-Thieme.
\newblock Bpr: Bayesian personalized ranking from implicit feedback.
\newblock In {\em UAI}, pages 452--461, 2009.

\bibitem[\protect\citeauthoryear{Robinson \bgroup \em et al.\egroup
  }{2020}]{robinson2020contrastive}
Joshua Robinson, Ching-Yao Chuang, Suvrit Sra, and Stefanie Jegelka.
\newblock Contrastive learning with hard negative samples.
\newblock {\em arXiv preprint arXiv:2010.04592}, 2020.

\bibitem[\protect\citeauthoryear{Shen \bgroup \em et al.\egroup
  }{2014}]{shen2014learning}
Yelong Shen, Xiaodong He, Jianfeng Gao, Li~Deng, and Gr{\'e}goire Mesnil.
\newblock Learning semantic representations using convolutional neural networks
  for web search.
\newblock In {\em WWW}, pages 373--374, 2014.

\bibitem[\protect\citeauthoryear{Sun \bgroup \em et al.\egroup
  }{2019}]{sun2019bert4rec}
Fei Sun, Jun Liu, Jian Wu, Changhua Pei, Xiao Lin, Wenwu Ou, and Peng Jiang.
\newblock Bert4rec: Sequential recommendation with bidirectional encoder
  representations from transformer.
\newblock In {\em CIKM}, pages 1441--1450, 2019.

\bibitem[\protect\citeauthoryear{Wang \bgroup \em et al.\egroup
  }{2021}]{wang2021cross}
Jinpeng Wang, Jieming Zhu, and Xiuqiang He.
\newblock Cross-batch negative sampling for training two-tower recommenders.
\newblock In {\em SIGIR}, pages 1632--1636, 2021.

\bibitem[\protect\citeauthoryear{Wei \bgroup \em et al.\egroup
  }{2021}]{wei2021contrastive}
Yinwei Wei, Xiang Wang, Qi~Li, Liqiang Nie, Yan Li, Xuanping Li, and Tat-Seng
  Chua.
\newblock Contrastive learning for cold-start recommendation.
\newblock In {\em MM}, pages 5382--5390, 2021.

\bibitem[\protect\citeauthoryear{Wu \bgroup \em et al.\egroup
  }{2019}]{wu2019npa}
Chuhan Wu, Fangzhao Wu, Mingxiao An, Jianqiang Huang, Yongfeng Huang, and Xing
  Xie.
\newblock Npa: Neural news recommendation with personalized attention.
\newblock In {\em KDD}, pages 2576--2584, 2019.

\bibitem[\protect\citeauthoryear{Xiong \bgroup \em et al.\egroup
  }{2020}]{xiong2020approximate}
Lee Xiong, Chenyan Xiong, Ye~Li, Kwok-Fung Tang, Jialin Liu, Paul Bennett,
  Junaid Ahmed, and Arnold Overwijk.
\newblock Approximate nearest neighbor negative contrastive learning for dense
  text retrieval.
\newblock {\em arXiv preprint arXiv:2007.00808}, 2020.

\bibitem[\protect\citeauthoryear{Yi \bgroup \em et al.\egroup
  }{2019}]{yi2019sampling}
Xinyang Yi, Ji~Yang, Lichan Hong, Derek~Zhiyuan Cheng, Lukasz Heldt, Aditee
  Kumthekar, Zhe Zhao, Li~Wei, and Ed~Chi.
\newblock Sampling-bias-corrected neural modeling for large corpus item
  recommendations.
\newblock In {\em Recsys}, pages 269--277, 2019.

\bibitem[\protect\citeauthoryear{Zhou \bgroup \em et al.\egroup
  }{2020}]{zhou2020s3}
Kun Zhou, Hui Wang, Wayne~Xin Zhao, Yutao Zhu, Sirui Wang, Fuzheng Zhang,
  Zhongyuan Wang, and Ji-Rong Wen.
\newblock S3-rec: Self-supervised learning for sequential recommendation with
  mutual information maximization.
\newblock In {\em CIKM}, pages 1893--1902, 2020.

\end{thebibliography}
